%% file: root.tex
\documentclass[letterpaper, 10 pt, conference]{ieeeconf}  % Comment this line out if you need a4paper

\IEEEoverridecommandlockouts                              % This command is only needed if 
\usepackage{times,amsmath,amsfonts,stmaryrd,multirow,subfigure,epsfig,enumerate}
\usepackage{breakurl}
\usepackage{flushend}
\usepackage{cite}
\usepackage{array}
\usepackage{color}
\usepackage{pdfsync}
\usepackage{textcomp}
\usepackage{graphicx}
\usepackage{algorithm}
\usepackage{algcompatible}
\usepackage{textcomp}
\usepackage{xfrac}
\usepackage{subfiles}
\usepackage{kotex}
\usepackage{hyperref}
\usepackage[dvipsnames]{xcolor}

\hypersetup{colorlinks=true,citecolor=blue,linkcolor=blue,linktocpage=true}

\title{\LARGE \bf
Visually Grounding Language Instruction for \\History-Dependent Manipulation
}

\author{Hyemin Ahn$^{1*}$, Obin Kwon$^{2*}$, Kyungdo Kim$^{3}$, Jaeyeon Jeong$^{2}$, Howoong Jun$^{4}$, Hongjung Lee$^{4}$,\\ Dongheui Lee$^{1, 5}$, Songhwai Oh$^{2}$% <-this % stops a space
\thanks{*These authors equally contributed to this work.}% <-this % stops a space
\thanks{$^{1}$Institute of Robotics and Mechatronics, German Aerospace Center (DLR), Wessling, Germany (e-mail: hyemin.ahn@dlr.de). $^{2}$Department of Electrical and Computer Enginnering and ASRI, Seoul National University, Seoul, Korea (e-mail: \{obin.kwon, jaeyeon.jeong\}@rllab.snu.ac.kr; songhwai@snu.ac.kr). $^{3}$Transdisciplinary Institute of Medicine \& Advanced Technology, Seoul National University Hopital, Seoul, Korea (e-mail: kimkyungdo08@gmail.com). $^{4}$Graduate School of Artificial Intelligence (GSAI) and ASRI, Seoul National University, Seoul, Korea (e-mail: \{howoong.jun, hongjung.lee\}@rllab.snu.ac.kr).$^{5}$Human-centered Assistive Robotics, Technical University of Munich, Munich, Germany (e-mail: dhlee@tum.de).}%
\thanks{This work was supported by Institute of Information \& Communications Technology Planning \& Evaluation (IITP) grant funded by the Korea government (MSIT) (No. 2019-0-01190, [SW Star Lab] Robot Learning: Efficient, Safe, and Socially-Acceptable Machine Learning), and partially supported by the Helmholtz Association.}%
}

\begin{document}
\sloppy

\maketitle
\thispagestyle{empty}
\pagestyle{empty}

%%%%%%%%%%%%%%%%%%%%%%%%%%%%%%%%%%%%%%%%%%%%%%%%%%%%%%%%%%%%%%%%%%%%%%%%%%%%%%%%
\begin{abstract}
This paper emphasizes the importance of a robot's ability to refer to its task history, especially when it executes a series of pick-and-place manipulations by following language instructions given one by one. 
The advantage of referring to the manipulation history can be categorized into two folds: (1) the language instructions omitting details but using expressions referring to the past can be interpreted, and (2) the visual information of objects occluded by previous manipulations can be inferred.
For this, we introduce a history-dependent manipulation task which objective is to visually ground a series of language instructions for proper pick-and-place manipulations by referring to the past.
We also suggest a relevant dataset and model which can be a baseline, and show that our model trained with the proposed dataset can also be applied to the real world based on the CycleGAN.
Our dataset and code are publicly available on the project website: \textcolor{MidnightBlue}{\url{https://sites.google.com/view/history-dependent-manipulation}}.
% Our dataset and code would be publicly available after the paper decision.
\end{abstract}

%%%%%%%%%%%%%%%%%%%%%%%%%%%%%%%%%%%%%%%%%%%%%%%%%%%%%%%%%%%%%%%%%%%%%%%%%%%%%%%%
\section{Introduction}

\begin{figure}[t]
\begin{center}
    \includegraphics[width=0.95\linewidth]{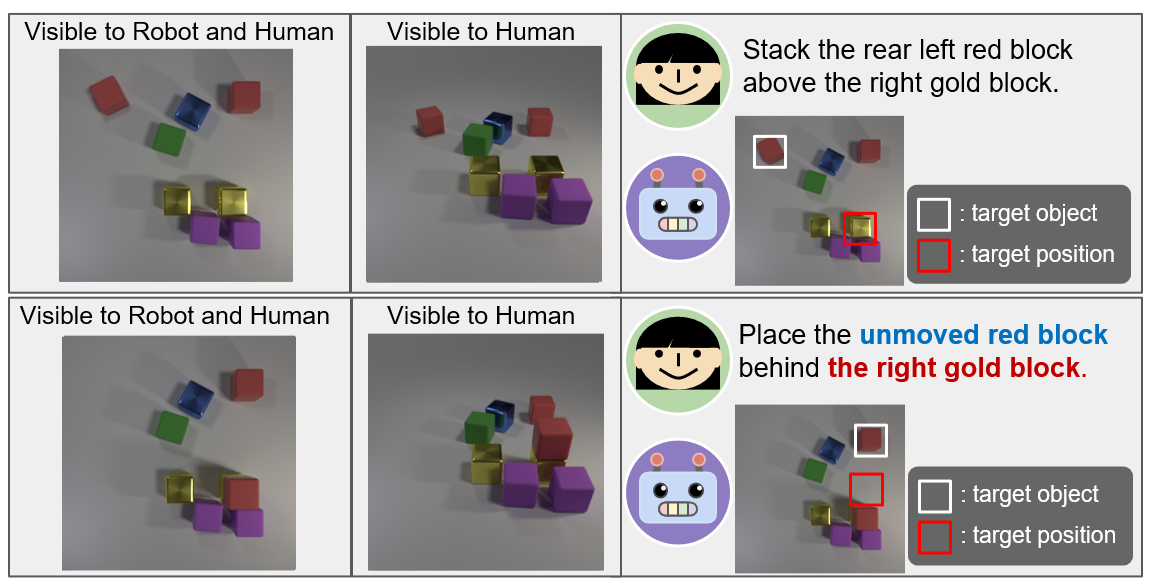}
\end{center}
\caption{An example scenario of a history-dependent manipulation task, where the robot is required to visually ground the language instruction by referring to its task history. By referring to the past, the robot can understand the human language that assumes the robot's ability to recall the past (blue fonts), and can infer the visually occluded information (red fonts).}
\label{fig:overview}
\vspace{-8mm}
\end{figure}

In this paper, our objective is to validate the benefits of the history-aware robots in manipulation, when the robot has to continuously receive language instructions one by one to perform a series of pick-and-place tasks.
To solve this problem, we propose a task of \textit{history-dependent manipulation}, whose example is shown in Figure~\ref{fig:overview}.
Given a shared workspace with rubber and metal blocks scattered around, the task of our robot is to manipulate the blocks one by one by following human language instructions to build a structure.
While the human can observe the workspace from multiple perspectives, the robot can only observe the workspace with a single RGB camera at a fixed position. 

In this setting, only relying on the current language and image inputs might be insufficient for robots to understand and execute the given task due to following reasons.
First, the problem would occur if we verbally instruct the robot by omitting details but using expressions referring to the past. 
The language expression in blue fonts in Figure~\ref{fig:overview} shows this case. 
After the human instructs the robot to manipulate `rear left red block', she orders to manipulate `\textit{unmoved} red block'. 
To interpret this, the robot needs to recall that the previously manipulated object was the `rear left red block', and understand what `\textit{unmoved} red block' refers to.

Second, the problem would occur if there is an occlusion from the robot's viewpoint after stacking or closely arranging objects.
Then, relying on the current visual observation would not be enough for robots to perform the task. 
The language expression with red fonts in Figure~\ref{fig:overview} shows an example of this case.
After the human instructs the robot to stack the rear left red block above the right gold block, the gold block becomes invisible to the robot at the next stage. 
However, since the right gold block is still visible from a human's perspective, one can instruct the robot to place the next target object behind the occluded gold block. 
Although the right gold block is invisible to the robot, this problem will be solved if the robot can recall the visual information obtained from the previous operation.

% For example, when a human instructs a robot as `Place the red block next to \textit{the block that you moved just before}', the robot needs to recall which block is the previously manipulated one.
%
% This can happen because of the human's common but unconscious assumption that others would understand the past and interpret the given utterance based on it.
%
% Second, if an occlusion occurs from the robot's viewpoint after stacking or closely arranging objects, relying on the current visual observation would not be enough for robots to perform the task. 
%

To solve these two problems, our robot needs to refer to its task history (1) for understanding the language expression assuming the robot's ability to recall the past, and (2) for overcoming the occlusion in its visual observation.
For these challenges, we propose a dataset of various scenarios of the \textit{history-dependent manipulation} task, and a deep neural network-based model to suggest a baseline for relevant future works.
Our dataset provides various tasks of history-dependent manipulations, where each task is composed of a series of pick-and-place operations for building a structure with given blocks. 
For each pick-and-place operation, a set of relevant instructions and RGB images observing the workspace are provided. 
The proposed model takes an RGB image and language instruction as inputs, and generates two heatmaps that estimate target object positions before and after the manipulation.
After each manipulation, a history vector that can abstract the obtained image, language, and time information is stored in the model's external memory. 
To visually ground the current language instruction, the stored history vectors are employed to perceive the current situation depending on the relevant manipulation history.

\section{Related Work}  
\paragraph{History-Dependent Instruction Following}
Instruction following for navigation 
\cite{rel_no_history_navi_XWORLD, rel_no_history_navi_XWORLD2D,rel_no_history_navi_chaplot18,rel_no_history_navi_EQA,rel_no_history_navi_reverie,rel_no_history_navi_vln,rel_no_history_navi_blukis20,rel_no_history_navi_blukis18_corl,rel_no_history_navi_blukis18_rss,rel_imp_history_navi_behavioral,rel_imp_history_navi_R4R,rel_imp_history_navi_RxR,rel_imp_history_navi_R6R,rel_imp_history_navi_touchdown,rel_imp_history_navi_RUN,rel_imp_history_navi_streetnav,rel_imp_history_navi_arramon,rel_imp_history_navi_vdn,rel_imp_history_navi_vdn_model,rel_imp_history_navi_hanna,rel_imp_history_navi_justask,rel_imp_history_navimani_alfred, rel_imp_history_navimani_lanichai,rel_exp_history_navimani_cerealbar,rel_exp_history_navi_robotslang}
or manipulation 
\cite{rel_no_history_mani_paxton19, rel_no_history_mani_Stepputtis20, rel_no_history_mani_Hatori18, rel_no_history_mani_bisk18,rel_no_history_mani_ingress,rel_no_history_mani_lynch21,rel_no_history_mani_bisk16, rel_imp_history_navimani_alfred, rel_imp_history_navimani_lanichai,rel_exp_history_navimani_cerealbar,rel_exp_history_navimani_minecraft} 
requires robots to iteratively interact with the environment.
The robot needs to remember what it has done so far to understand the current environment correctly.
% naturally requires an understanding of historical information.
% %
% Since the robot iteratively interacts with the environment, it needs to remember what it has done so far to understand the current state of the environment correctly.
%
% Many of the existing researches \cite{rel_no_history_navi_XWORLD, rel_no_history_navi_XWORLD2D,rel_no_history_navi_chaplot18,rel_no_history_navi_EQA,rel_no_history_navi_reverie,rel_no_history_navi_vln,rel_no_history_navi_blukis20,rel_no_history_navi_blukis18_corl,rel_no_history_navi_blukis18_rss,rel_no_history_mani_paxton19, rel_no_history_mani_Stepputtis20, rel_no_history_mani_Hatori18,rel_no_history_mani_bisk18,rel_no_history_mani_ingress,rel_no_history_mani_lynch21,rel_no_history_mani_bisk16} developed frameworks which can interact with the world, following the given instructions.
% %
However, their instructions are history-independent (e.g., What color is the chair in the bedroom?), explicitly explaining the target without temporal deixis.
%
%The history-independent instructions can be fully interpreted without any ambiguity at any timestep (or the ambiguity cannot be resolved even with the history information).
%
They focus on mapping an instruction to a series of robot actions, without having to recall the past to understand the current instruction.
% They focused on mapping instruction to the series of robot actions, with the agent aware of its actions so far and what to do next.

Most of the history-dependent expressions in the existing datasets are simple sequencing expressions in the navigation task \cite{rel_imp_history_navi_behavioral, rel_imp_history_navi_R4R, rel_imp_history_navi_RxR,rel_imp_history_navi_R6R,rel_imp_history_navi_touchdown,rel_imp_history_navi_RUN,rel_imp_history_navi_streetnav,rel_imp_history_navi_arramon,rel_imp_history_navi_vdn,rel_imp_history_navi_vdn_model,rel_imp_history_navi_hanna,rel_imp_history_navi_justask, rel_imp_history_navimani_lanichai}, such as `Go straight and turn right at the \textit{second} intersection'.
Dialogue-guided navigation dataset RobotSlang \cite{rel_exp_history_navi_robotslang} contains few complex history-dependent expressions (below 0.1\%, e.g., `Go back to the brown wall \textit{you were at before}'). Household task instruction dataset ALFRED \cite{rel_imp_history_navimani_alfred} also contains some history-dependent expressions, referring to the objects previously seen but not present in the robot view.
Compared to the previous tasks, our history-dependent manipulation task focuses more on the advanced understanding of history-dependent instructions. We explicitly define the difficulties when the language instructions can only be interpreted by referring to the task history, and suggest a method to resolve such difficulties.
%
% In addition, there are several works related to the cooking scenarios, where the robot should be aware what it have done so far, to determine what to do next. \textcolor{blue}{Obin: please add reference here.}
% In, a robot is also required to remember th ehas to navigate through a path specified by given language instructions. 
%It has to remember where it has passed to track how much the agent followed the instruction. 
%Vision-and-dialog navigation requires a robot to keep asking a human how to navigate until reaching a target, and a human would respond to the robot. 

% (혜민) 이부분은 맡기겠습니다 저번에 냈던거                     https://www.overleaf.com/3353552621dspmmsznkrxn 같이 참고해서 써주세욥! 릴레이티드 워크 총 한 장이 만들어질 정도로 써주세요.

% Many of the robot tasks that have been studied before requires history-understanding.
% Compared to these tasks, our history-dependent manipulation task is related to the instruction following where a robot needs to understand the human utterance and perform a goal task. 

Of course, there exist related works tackling the similar problem of understanding history-dependent instructions.
\cite{tellmedave} suggests a system to ground a given language instruction
% to the proper manipulation type 
depending on the environment as well as the task context.  
One of the main challenges in \cite{tellmedave} is to understand anaphoric references based on the history (i.e., Pick up the \textit{snack}, and microwave \textit{it}). 
In \cite{tgg_mit}, authors propose a probabilistic model, which can accumulate the knowledge based on the past visual and language information, and employ that knowledge to ground the current language instruction.
%
% Compared to these works, our history-dependent manipulation task is more complicated and focuses more on robot's ability to refer the task history.
% extends these works to the more complicated task.

These works focus on understanding the context that happened one or two steps before the current instruction. 
Compared to this, instructions in our history-dependent manipulation task sometimes refer to the past far from the current step. %
For example, our language input can refer to ``the block you moved at the \textit{first time}", when the robot is conducting the \textit{sixth} pick-and-place operation.
%\DL{Is there somewhere giving the number, how many steps before?}->resolved by adding example here.
This makes our task more challenging since robots need to recall task history through multiple interactions. 
Furthermore, our study suggests that understanding the task history can also help robots to infer the visual information about occluded objects, which is another crucial aspect not covered by previous relevant studies.
% Recently, a task dataset named ALFRED has been proposed \cite{ALFRED}, and it is concurrent with our work. 
% In this task, a robot needs to navigate through the environment as well as manipulate objects based on the given step-by-step instructions (e.g., Move to the wooden shelf. Pick up the salt from the shelf. Bring the salt to the counter right of the fridge.) 
% In this dataset, the instructions are clearly stated without anaphoric references, however, the referred objects are sometimes invisible from the robot's perspective. Therefore, a robot needs to remember its own observation and action history to understand the instruction.
% Compared to this task, our history-dependent manipulation is more focused on the advanced understanding of history-dependent instructions. We explicitly define the difficulties when the language instructions can only be interpreted by referring to the task history, and suggests a method to resolve such difficulties. 
% While VLN tasks focus on the navigation to the referred place and object class, the history dependent manipulation focuses on resolving ambiguity using past information. In our proposed task, a robot has to distinguish the referred instance among the similar class of objects using spatio-temporal relationship between each elements.

\paragraph{Visually Grounding Referring Expressions}
Since our robot needs to ground the given language instructions in the RGB image, our history-dependent manipulation task has a similarity with several studies in the computer vision research field, which comprehend the referring expressions that describe a target object in the input image \cite{rel_det_1, rel_det_2, rel_seg_1, rel_seg_2, rel_seg_3, rel_mattnet, rel_cvpr2020}.
%\cite{johnson2017clevr,clevr_dialog,refcoco,google_refexp, refclef}
%
Related studies can be categorized by whether the proposed model finds the target based on a detection box \cite{rel_det_1, rel_det_2}, segmentation map \cite{rel_seg_1, rel_seg_2, rel_seg_3}, or both \cite{rel_mattnet, rel_cvpr2020}.

This task of visually grounding referring expressions is also related to object manipulation \cite{rel_mani_dataset, rel_mani_brown}, for enabling robots to find the target object to manipulate.
\cite{rel_mani_dataset} proposes a dataset that can be used to train a robot picking up the target object referred to by language instructions.
%
% \cite{rel_mani_rl} suggested using a reinforcement learning based agent which can move the target object to the desired location in the 2-dimensional workspace, by mapping the visual observation and given language command to the corresponding action.
% %
\cite{rel_mani_brown} develops a system that can distinguish the target object referred to by language instruction when multiple objects belonging to the same class are given.
These studies are similar to ours in that their models can retrieve the object referred to by a given language expression.
However, since they do not assume that a human can instruct a robot over several times, their goal does not include enabling robots to understand the \textit{historical} information accrued through the past.

There exist similar tasks as ours in natural language processing literature which studies a two-player collaborative scenario to collect the cards \cite{rel_exp_history_navimani_cerealbar} or build an architecture \cite{rel_exp_history_navimani_minecraft} in a 3D environment.
In \cite{rel_exp_history_navimani_cerealbar,rel_exp_history_navimani_minecraft}, the players use multiple history-dependent expressions to complete the given task. However, their focus is on language expression understanding, assuming the agent has full observability. 
In this paper, we study more practical issue that arises from using a real robot with partial observability.

\section{Dataset} \label{sec:data}
\begin{figure*}[t]
    \begin{center}
        \includegraphics[width=0.85\linewidth]{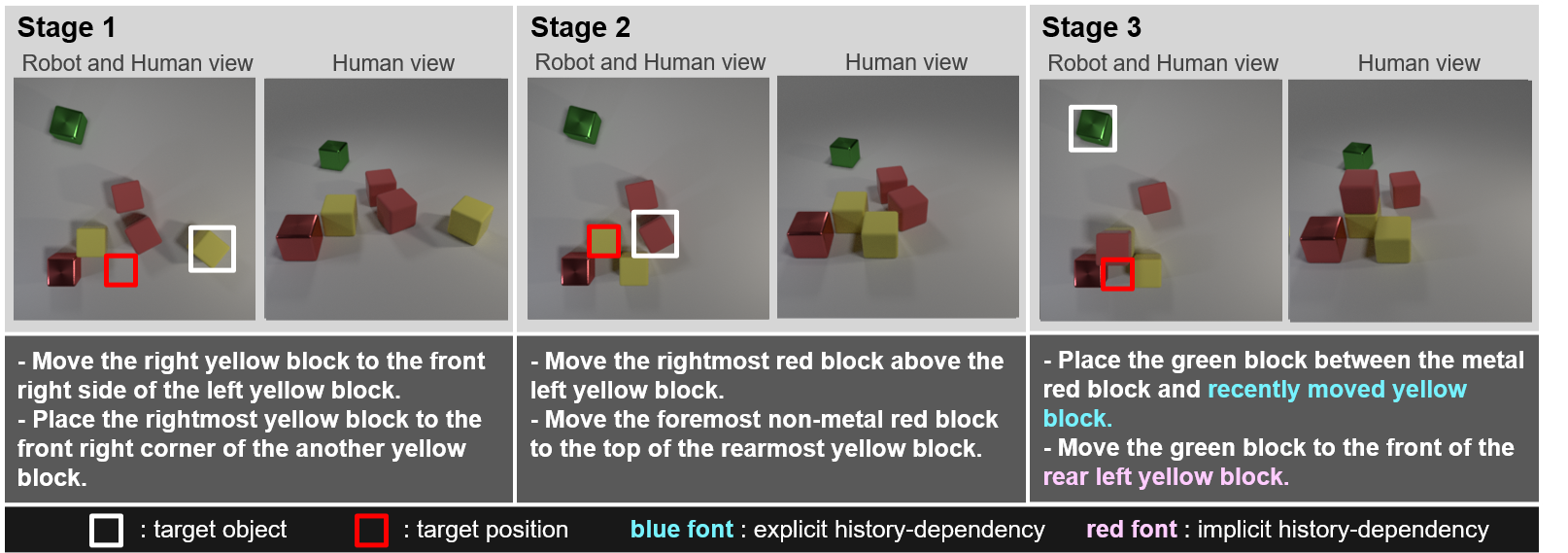}
    \end{center}%
\caption{The example of the proposed dataset for a history-dependent manipulation task consisting of three pick-and-place operations. The first row shows images from perspectives of robot and human, the second row shows the example language instructions. Ground truth positions of the target object before and after the manipulation are annotated with white and red boxes.}
\label{fig:dataset}
\vspace{-5mm}
\end{figure*}

As shown in Figure~\ref{fig:dataset}, our history-dependent manipulation task consists of several pick-and-place operations, which are instructed by a real human language.
Each task starts with a shared workspace with several blocks of various colors and materials. 
A human gives a set of language instructions for a robot to move given blocks one by one for building a specific structure.
The main objective of the robot is to estimate the position of the target block before and after executing the given instructions. 
During the task, it is assumed that the robot observes the workspace vertically from above using an RGB camera.
The human can observe the workspace from multiple perspectives, but one provides the instruction based on the observation from the front side (i.e., `human view' images in Figure \ref{fig:dataset}).
Since the robot's camera would be fixed at a predefined position, there can be blocks that are invisible or partially visible to the robot due to occlusion.

For each pick-and-place operation, our dataset provides synthetic RGB images of the workspace from two viewpoints, a set of language instructions from real humans, bounding boxes, and heatmaps showing the target object's position before and after the manipulation.
With our simulator based on the code from \cite{johnson2017clevr}, we collect images reflecting this task scenario.
After spawning blocks with different colors and materials, our simulator for image generation randomly chooses which block to move and where to put it.
The target position of the block can be among the predefined positions (i.e., center, front left, rear right) inside the workspace, or the position nearby other blocks. 

After collecting synthetic images as shown in Figure~\ref{fig:dataset}, we recruit six human participants to annotate each pick-and-place operation with language instructions.
The tasks are evenly distributed to the participants so that at least two people can annotate each task.
%
% For each image, one to four language instructions are collected, and each history-dependent manipulation task consists of 3 to 6 pick-and-place operations. 
%
The style of the collected language instructions is different from each human participant, which results in the challenging instruction dataset.
In total, the dataset comprises 300 scenarios of history-dependent manipulation tasks (250 for training and 50 for the test), 1339$\times$2 images from the robot and human viewpoints for 1339 pick-and-place operations, and 4642 language instructions.

In our dataset, the collected language instructions can be categorized as follows: (1) one that requires the robot to recall how blocks moved by using explicit expressions such as `\textit{previous}', `\textit{that you moved before}', (2) another that refers to blocks that are invisible or partially visible from the robot's viewpoint, and (3) others that do not belong to the previous two categories.
We will call the first type of history-dependency in language instructions as \textit{explicit history-dependency} (blue fonts in Figure \ref{fig:dataset}), and the second type as \textit{implicit history-dependency} (red fonts in Figure \ref{fig:dataset}). 

\begin{figure*}[t]
\begin{center}
    \includegraphics[width=0.89\linewidth]{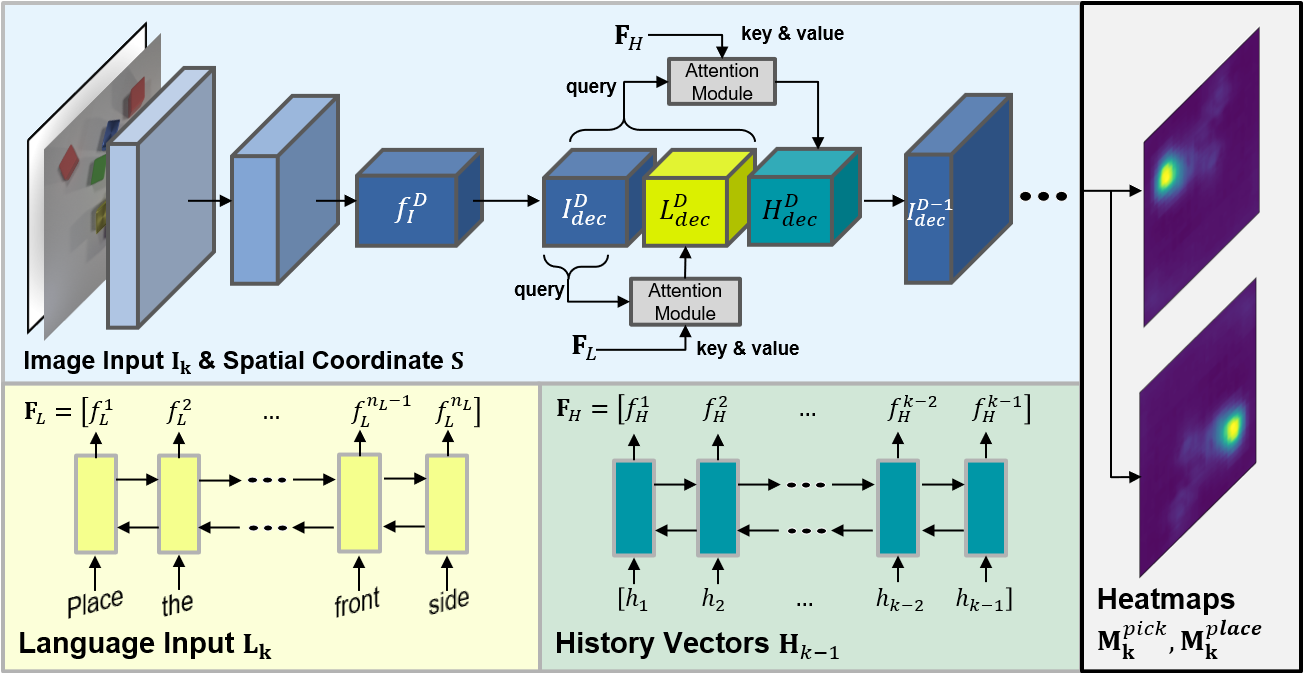}
\end{center}
\caption{Structure of the proposed model for the history-dependent manipulation task. The model consists of the hourglass network \cite{hourglass} (upper blue zone), bidirectional LSTM for language information (lower left yellow zone), and another bidirectional LSTM for history information (lower right green zone). In the bottom-up process, a set of spatial features is encoded from the input image and spatial coordinate. In the top-down process, the obtained language and history information are fused into the spatial features to obtain the desired heatmaps.}
\label{fig:model}
% \vspace{-5mm}
\end{figure*}

\section{Methodology}
\subsection{Overview}
In our history-dependent manipulation task, let us assume that a human instructs a robot for $n$ times. 
Then, let $\mathbf{I}_k \in \mathbb{R}^{w_I \times h_I \times 3}$ denote the $k$-th input RGB image, and $\mathbf{L}_k = \{ w_1, \ldots, w_{n_L} \}$ denote the $k$-th input language sentence with length $n_L$, where $w_i$ denotes an one-hot encoding vector representing the $i$-th word in the sentence.
In addition, let $\mathbf{H}_{k-1} = \{ h_1 \ldots h_{k-1} \}$ denote a set of history vectors accrued over $k-1$ times of pick-and-place operations. 
In our model, the visual, linguistic, and time information related to the $k$-th pick-and-place operation is abstracted into $h_k$, and stored in $\mathbf{H}_{k-1}$.

Let $\mathcal{F}$ denote our model. It takes the $\mathbf{I}_k$, $\mathbf{L}_k$, and $\mathbf{H}_{k-1}$ as inputs and results in two-dimensional positional heatmaps $\mathbf{M}^{pick}_k, \mathbf{M}^{place}_k \in \mathbb{R}^{w_M \times h_M}$, and one history vector $h_{k}$:
\begin{align}
    \mathbf{M}^{pick}_k, \mathbf{M}^{place}_k, h_{k} = \mathcal{F}(\mathbf{I}_k, \mathbf{L}_k, \mathbf{H}_{k-1})
    \nonumber
\end{align}
Here, $\mathbf{M}^{pick}_k$ and $\mathbf{M}^{place}_k$ are heatmaps indicating the confidence in the position of the target object to \textit{pick} and where to \textit{place} it. 
%
% We select to use heatmap-based network due to its high interpretability of obtained results, as shown in Section~\ref{sec:qual} and supplementary material.
%
Based on the obtained two heatmaps, the robot executes the pick-and-place operation based on the position where the value of each heatmap for pick/place is the highest. 
After the manipulation, the obtained history vector $h_{k}$ is stored to $\mathbf{H}_{k-1}$ for the next operation.

\subsection{Network Structure}
Figure~\ref{fig:model} shows the structure of the proposed  model $\mathcal{F}$, which is influenced by the network structure from \cite{it2p}.
The proposed $\mathcal{F}$ consists of (1) an hourglass network \cite{hourglass} to encode the image information as well as to decode all information into the desired heatmaps (upper blue zone), (2) a bidirectional LSTM to encode the language information (lower left yellow zone), and (3) a bidirectional LSTM to encode the history information (lower right green zone).
\paragraph{Image Information}

When $\mathbf{I}_k$ is given as an input, the \textit{spatial coordinates} $\mathbf{S}$, whose size is same as $\mathbf{I}_k$ and including the xy-positional information of each spatial location, is concatenated to the $\mathbf{I}_k$ first. 
The implementation of $\mathbf{S}$ is similar to the approach proposed in \cite{spatial_coordinate}.
With concatenated $\mathbf{I}_k$ and $\mathbf{S}$ as an input, the hourglass network \cite{hourglass} encodes a set of spatial features from high to low resolution based on its bottom-up process with max-pooling layers and residual modules \cite{resnet}.
Let $D$ denote the depth of the hourglass network, which implies that the bottom-up process is conducted for $D$ times, and Figure~\ref{fig:model} shows a model when $D=3$.
Let $\mathbf{F}_{I} = \{f_{I}^1 \ldots f_{I}^D \}$ denote our image information, which is a set of encoded spatial features from the all bottom-up processes. 
Here, $f_{I}^d \in \mathbb{R}^{w_d \times h_d \times l_d}$ denotes the spatial feature from the $d$-th bottom-up process. 
%Here, the channel dimension $l$ is same for all $f_{enc}$.
%
% After passing the last spatial feature with the lowest resolution $f_{enc}^D$ into the avg-pooling layer, we obtain an image feature vector $f_I$.
% %
The top-down process of the hourglass network starts from $f_{I}^D$, and it is described in the last paragraph of this section.

\paragraph{Language Information}
When the language sentence $\mathbf{L}_k=\{w_1 \ldots w_{n_L}\}$ is given as an input, all one-hot encoding word vectors $w_i$ are converted into the embedding representation $e_i$ with the trainable embedding matrix $E$, such that $e_i = E w_i$. 
When $\{e_1 \ldots e_{n_L} \}$ is given to the bidirectional LSTM, a set of hidden state vectors of the LSTM is obtained and used as our language information $\mathbf{F}_{L}=\{f_L^1,\ldots,f_L^{n_L}\}$.
It is used in the top-down process of the hourglass network, which fuses all information to obtain two desired heatmaps.
\paragraph{History and Time Information}
A set of history vectors $\mathbf{H}_{k-1} = \{h_1,\ldots h_{k-1}\}$ is also given to the bidirectional LSTM. Then, the obtained set of hidden state vectors of the LSTM is used as our history information $\mathbf{F}_H=\{f_H^1,\ldots,f_H^{k-1}\}$.
This history information $\mathbf{F}_H$ is fused with image and language information in the top-down process of the hourglass network.

In addition, we also encode the time information into a feature vector $f_T$, by projecting the one-hot encoded time index vector $T$ into the higher dimensional space. For example, if the human instructs the robot for the second time, the time indexes would be $T=[0, 1, 0, 0,\ldots]^T$, and $T$ is projected into the higher dimensional time feature vector $f_T$ by a matrix multiplication.

\paragraph{Attention Module}
Before describing the top-down process of the hourglass network, we would explain the attention module first, which is used in the top-down process to fuse image, language, and history information for obtaining desired heatmaps.
Let $\mathbf{X}=\{x_1,\ldots,x_N\}$ denote a set of feature vectors, where $x_i \in \mathbb{R}^{D_x}$.
Based on the attention mechanism from \cite{transformer}, the attention module determines where to attend on the information in $\mathbf{X}$.

Let $\mathbf{q} \in \mathbb{R}^{D_{qkv}}$ denote the given query vector, and denote $W_k, W_v \in \mathbb{R}^{D_{qkv} \times D_x}$ as matrices for projecting $x_i$ into key and value vectors as $W_k x_i, W_v x_i \in \mathbb{R}^{D_{qkv}}$. 
Based on this, a feature vector $\mathbf{f}_X$ representing $\mathbf{X}$ is obtained by a weighted sum of values, where the weight for the $i$-th value $W_v x_i$ is a softmax score of the $i$-th key $W_k x_i$, computed based on the dot product between $\mathbf{q}$ as follows:
\begin{align}
    \mathbf{f}_X
    =& \mathcal{A}(\mathbf{q}, \mathbf{X}; W_k, W_v) 
    = \sum_{i=1}^{N}a_i (W_v x_i),
    \nonumber
    \\
    a_i 
    =& \frac{\exp(\textrm{score}({\mathbf{q}, W_k x_i}))}{\sum_{j=1}^{N}\exp(\textrm{score}({\mathbf{q}, W_k x_j}))},\;     
    \textrm{score}(\mathbf{m}, \mathbf{n}) 
    = \mathbf{m}^T\mathbf{n}/\tau
    \nonumber
\end{align}
Here, a scalar $\tau$ represents a temperature parameter.

\paragraph{Top-down Process for Heatmap Generation}

After obtaining the image, language, and history information, our model decodes all into the desired heatmaps based on the top-down process with upsampling and skip connection.
The top-down process starts from $f_{I}^D$, the last spatial feature from the bottom-up process.

First, we obtain $I_{dec}^D$ by passing $f_I^D$ to a single residual module.
Let $I_{dec}^D(i, j)$ denote a feature vector from $I_{dec}^D$, at a spatial location $(i, j)$.
We fuse the language information $\mathbf{F}_L$ to each spatial location of $I_{dec}^D$ based on the attention module.
To do this, $I_{dec}^D(i, j)$ is used as a query and $\mathbf{F}_L$ is used as keys and values with matrices $W_k^L$ and $W_v^L$, such that:
\begin{equation}
    L_{dec}^D(i, j)=\mathcal{A}(I_{dec}^D(i, j), \mathbf{F}_L; W_k^L, W_v^L)
    \nonumber
\end{equation}

\input{table1.tex}
\begin{figure}[t]
\begin{center}
    \includegraphics[width=\linewidth]{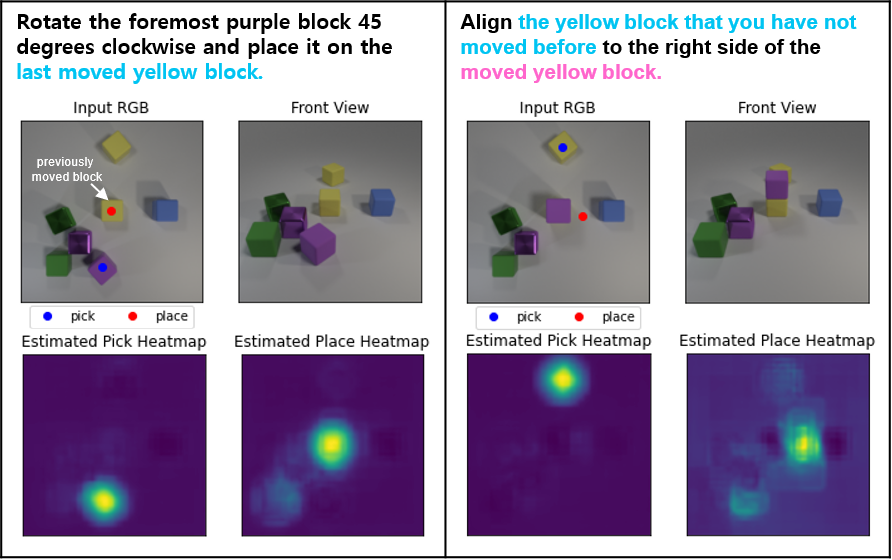}
\end{center}
\caption{Generated heatmap results from a history-dependent task in our test dataset. In input RGB images, the blue and red dots denote positions where the values of the estimated heatmaps are the highest.}
\label{fig:qual_result}
\vspace{-3mm}
\end{figure}

Then, every $L_{dec}^D(i, j)$ is concatenated with $I_{dec}^D(i, j)$, and used as a query to fuse the history information $\mathbf{F}_H$ for each spatial location $(i, j)$: 
\begin{equation}
    H_{dec}^D(i, j)=\mathcal{A}([I_{dec}^D(i, j); L_{dec}^D(i, j)], \mathbf{F}_H; W_k^H, W_v^H)
    \nonumber
\end{equation}
After this process, $I_{dec}^D$, $L_{dec}^D$, and $H_{dec}^D$ are concatenated again.
The concatenated result passes the residual module as well as the skip connection with $f_I^{D-1}$, and $I_{dec}^{D-1}$ is obtained. 
This process is repeated until the $I_{dec}$ has a desired spatial dimension.
The final $I_{dec}$ from the top-down process is converted into the desired heatmaps $\mathbf{M}_k^{pick}$ and $\mathbf{M}_k^{place}$, after it passes one residual module and several convolutional layers. 
The history vector $h_k$ summarizing the image, language, and time information of this pick-and-place operation is obtained as 0f, where $avgpool$ denotes the average pooling layer.

\subsection{Network Training}
Our dataset consists of 300 history-dependent manipulation tasks, where 250 tasks are for training, and 50 tasks are for the test.
Note that a task consists of a series of pick-and-place operations, and our dataset provides one to four language instructions describing each pick-and-place operation.
Therefore, when training the model over one task, one sentence is randomly sampled as an input for each pick-and-place. 
For the data augmentation, we randomly flip images horizontally and change the sentence to correspond to the flipped image.
After the model results in a series of heatmaps,
%$\{(\mathbf{M}_k^{pick}, \mathbf{M}_k^{place})\}_{k=1...n}$,
a mean square error is calculated based on the ground truth heatmaps and it is minimized by Adam optimizer \cite{adam}.
Note that the whole model is trained in an end-to-end way.

\section{Experiments}
\subsection{Qualitative Result}
Figure~\ref{fig:qual_result} shows an example of the qualitative result from the proposed model and test dataset.
Here, the blue and red dots in input RGB images denote the positions where the values of the estimated heatmaps are maximum.
In language instructions, the blue fonts show the \textit{explicitly history-dependent} expressions which include language expressions explicitly referring to the past (i.e., `\textit{last}', `\textit{moved before}'). And the red fonts show the \textit{implicitly history-dependent} expressions which refer to the occluded object.
The first instruction orders the robot to stack the foremost purple block above the `last moved yellow block'.
Note that the center yellow block in the first input RGB image is the one that moved before.
The obtained heatmaps show that the model successfully estimated which yellow block is the previous one.
In the second instruction, the human refers to the remained yellow block as `the yellow block that you have not moved before'.
Also, the center yellow block is referred to as `moved yellow block', but it is not visible to the robot due to the occlusion.
Generated heatmaps for this operation show that the model successfully estimates the position of the target object and where to place it, even if there are explicit/implicit history-dependent expressions in the same language sentence. 

\subsection{Quantitative Result}
\paragraph{Ablation Study}
To validate the effectiveness of exploiting the history information in our task, we compare the performance between the proposed models with and without the history information $\mathbf{F}_H$.
When our model does not exploit the history information, it neglects $\mathbf{F}_H$ in its top-down process, so that only the image and language information is employed to obtain the desired heatmap.
To measure the model performance, we calculate the accuracy (\%) of pick-and-place operations consisting of the history-dependent manipulation tasks in our test dataset.
The accuracy for a single pick-and-place operation is measured by randomly sampling one sentence in test data as an input.
Note that one to four language instructions can exist for describing one pick-and-place operation.
The result is defined as successful when the distance between the predicted pick/place positions and ground truth positions is less than 15 pixels, which is about the half size of the block when the image size is 256$\times$256.

Table~\ref{tab:quan_result} shows the comparison results between proposed models with and without the history information.
Results are classified based on the type of history-dependency in the input language instructions. 
% About definitions of types of history-dependency, please refer the Section~\ref{sec:data}.  
%
When the instruction requires the robot to recall its task history, the performances are higher from the proposed model with history information.
However, when there is an implicit history dependency, it is shown that the performance gap is lower. We expect this is because the model without the history information could also work when the occluded object is `partially' visible.
But still, we claim that the result implies that the history-aware robot can solve a task better when a human explicitly/implicitly requires the robot to employ the task history information.

\paragraph{Comparison with MAttNet \cite{rel_mattnet}}
\input{table2}
In addition to the ablation study, we compare the performance of our model with MAttNet \cite{rel_mattnet}, which is one of the representative studies of visually grounding referring expressions. 
We train two separate MAttNets for each pick/place operation.
Note that the MAttNet does not have the ability to refer to its task history, but we add this to our comparison study to check how much our model is adequate to be a baseline.

\begin{figure}[t]
\begin{center}
    \includegraphics[width=0.9\linewidth]{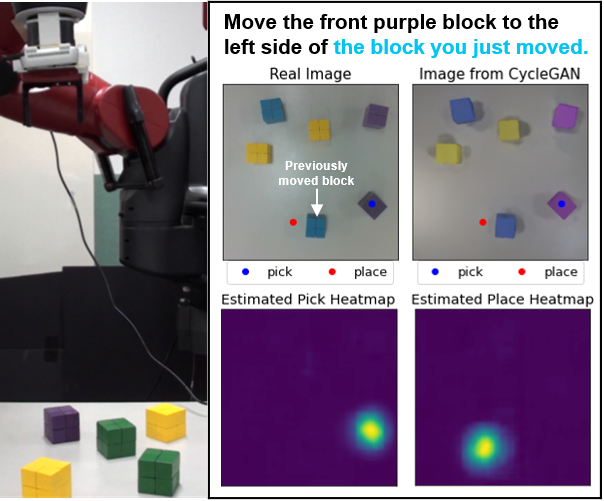}
\end{center}
\caption{An example of applying the trained model to the real world. Input language instruction is grounded to the image transferred by CycleGAN \cite{CycleGAN}. In input images, blue and red dots denote the positions where the values of the estimated heatmaps are the highest. More real world demonstrations are included in our supplementary video.}
\label{fig:real_result}
\vspace{-5mm}
\end{figure}

Table~\ref{tab:mattnet} shows the pick-and-place accuracy (\%) of the proposed model and MAttNet, where the accuracy is measured with the same method used in the ablation study.
MAttNet outperforms when there is no history-dependent expression in the input language, since it is specialized in the task of visually grounding referring expression.
It first detects the candidate regions from the image and classifies which region is referred to by the given sentence. This makes it easier for MAttNet to solve the problem than heatmap-based approach like us, since the candidates are explicitly defined in advance. 
But note that our MAttNet implementation based on the official code has 52.7M network parameters, and takes 64.4 hours for training, with Nvidia Titan X GPU, 128GB RAM, Intel i7 CPU.
Compared to this, ours has 39.6M network parameters, and takes 2 hours for training, with Nvidia RTX 2060 SUPER GPU, 32GB RAM, Intel i7 CPU.

The result shows that our model performs better when there is explicit or implicit history dependency. 
% Especially, it is shown that the performance gap between the proposed model and MAttNet is the highest when there is an implicit history dependency.
%
This shows that exploiting the history information makes our model more effective in solving the history-dependent manipulation task.
Based on this, we claim that the our model can be a comparable baseline for future studies relevant to the history-dependent manipulation task.

\subsection{Real World Demonstration}
We also show that our model can be also applied to the real world.
As shown in Figure~\ref{fig:real_result}, we attach a RealSense D435 to the Baxter robot's right hand and place its arm at a fixed position to observe the workspace.
To bridge the gap between RGB images from the real world and the dataset, we employ CycleGAN \cite{CycleGAN}, which can transfer images to different domains.
To train CycleGAN, we collect 354 RGB images of real-world workspace and employ 660 images from our training dataset.
As shown in Figure~\ref{fig:real_result}, the trained CycleGAN transforms the real-world RGB images to look like the ones from our dataset, so that our model can execute the task based on the transferred image.

With the transformed RGB image and the given language instruction, our model infers two heatmaps of pick-and-place.
Then, we choose two pixel positions with the highest pick/place heatmap values.
By using the depth image of the calibrated RealSense camera, a robot roughly estimates the real-world xyz-positions of the chosen pixel positions, and uses them for manipulation.
We use the built-in inverse-kinematics solver of the Baxter robot to generate its proper arm trajectory.
Here, we also leverage the inner-wrist RGB camera in the Baxter robot.
This camera is used to adjust the gripper position, after reaching to the roughly estimated xyz-position from the RealSense camera.
After reaching to the estimated xyz-position, we calculate the contours of the target block from the RGB image and re-estimate the precise pose of the block. 
For more real world demonstration cases, please refer to our supplementary video.

% \section{Limitation and Future Work}

% There exist several limitations in our work. First, all images are synthetic and based only on cubes. Second, robot behavior is only limited to pick-and-place.

\section{Conclusion}
In this paper, we define a task of \textit{history-dependent manipulation}, which aims to enable a robot to refer to its task history when executing a series of pick-and-place operations instructed by language.
In this task, two main challenges can arise: (1) the language instructions referring to the past can be given, and (2) the objects occluded after pick-and-place operations need to be inferred.
To solve this, we propose a dataset and neural network-based model, and apply the model to the real world to suggest the scalability of our work.
However, since we simplify several factors to focus on the scenario when understanding the task history becomes the most crucial factor to successfully perform a task, there exist limitations such as (1) images are synthetic, and (2) the robot behavior is limited to pick-and-place. 
But still, our results show that the proposed end-to-end-trainable model can successfully learn how to understand the history by abstracting the visual, language, and time information from the previous pick-and-place operations.
Our work will provide a reasonable baseline for studies of history-aware manipulation robots, and improve the chance of realizing long-term human-robot collaboration in the future.
% based on the attention mechanism determining which part of the information needs to be focused on.
% exploit the history information which is built
% Based on the CycleGAN \cite{CycleGAN}, our model can also be applied to the real world, suggesting the scalability of our work.

\bibliography{main}
\bibliographystyle{IEEEtran}
\end{document}

%% file: table1.tex
\begin{table}[t]
\centering
\caption{Pick-and-place accuracy (\%) comparison between models with and without history information $\mathbf{F}_H$}
\begin{tabular}{|l|c|c|c|} 
\hline
                              & \multicolumn{3}{c|}{History Dependency}        \\ 
\cline{2-4}
                              & None          & Explicit      & Implicit       \\ 
\hline
Proposed with $\mathbf{F}_H$ & \textbf{73.0} & \textbf{62.2} & \textbf{68.3}  \\ 
\hline
Proposed w/o $\mathbf{F}_H$  & 71.3          & 54.9          & 66.7           \\
\hline
\end{tabular}
\label{tab:quan_result}
% \vspace{-3mm}
\end{table}

%% file: table2.tex
\begin{table}
\centering
\caption{Pick-and-place accuracy (\%) of our model and MAttNet \cite{rel_mattnet}}
\begin{tabular}{|l|c|c|c|} 
\hline
         & \multicolumn{3}{c|}{History Dependency}        \\ 
\cline{2-4}
         & None          & Explicit      & Implicit       \\ 
\hline
Proposed & 73.0          & \textbf{62.2} & \textbf{68.3}  \\ 
\hline
MAttNet \cite{rel_mattnet}  & \textbf{81.1} & 61.6          & 56.5           \\
\hline
\end{tabular}
\label{tab:mattnet}
% \vspace{-7mm}
\end{table}